\pdfoutput=1
\documentclass[10pt]{article}
\usepackage[margin=1in]{geometry} % to change the page dimensions
\usepackage{enumerate, amssymb, amsfonts, latexsym}
\usepackage{amscd, amsmath, amsthm}
\usepackage{graphicx}
\usepackage{bm}
\usepackage{mathtools}
\usepackage{caption}
\usepackage{subcaption}
\theoremstyle{plain}
\newtheorem{theorem}{Theorem}[section]
\newtheorem{lemma}[theorem]{Lemma}

\theoremstyle{definition}
\newtheorem{definition}[theorem]{Definition}

\usepackage{hyperref}
\usepackage{wrapfig}
\hypersetup{colorlinks,    citecolor=black,
    filecolor=black,
    linkcolor=black,
    urlcolor=red}
\DeclareMathOperator*{\argmin}{\arg\!\min}

\DeclareMathOperator*{\E}{\mathbb{E}}
\begin{document}

\title{Learning Parameters for Weighted Matrix Completion via Empirical Estimation}
\author{Jason Jo \\ 
jjo@math.utexas.edu \\
Mathematics Department at the University of Texas at Austin
\\ 2515 Speedway,
Austin, Texas 78712}
\date{}
\maketitle

\begin{abstract}
Recently theoretical guarantees have been obtained 
for matrix completion in the non-uniform sampling regime. In particular, if the 
sampling distribution aligns with the underlying matrix's leverage scores, then 
with high probability nuclear norm minimization will exactly recover the low 
rank matrix. In this article, we analyze the scenario in which the non-uniform 
sampling distribution may or may not not align with the underlying matrix's 
leverage scores. Here we explore learning the parameters for weighted nuclear 
norm minimization in terms of the empirical sampling distribution. We 
provide a sufficiency condition for these learned weights which provide an exact 
recovery guarantee for weighted nuclear norm minimization. It has been 
established that a specific choice of weights in terms of the true sampling 
distribution not only allows for weighted nuclear norm 
minimization to exactly recover the low rank matrix, but also allows for a quantifiable relaxation in the exact recovery conditions. In this article we extend this quantifiable relaxation in exact recovery conditions for a specific choice of weights defined analogously in terms of the empirical distribution as opposed to the true sampling distribution. To accomplish this we employ a concentration of measure bound and a large deviation bound. We also present numerical evidence for the healthy robustness of the weighted nuclear norm minimization algorithm to the choice of empirically learned weights. These numerical experiments show that for a variety of easily computable empirical weights, weighted nuclear norm minimization outperforms unweighted nuclear norm minimization in the non-uniform sampling regime.  
\end{abstract}

%\tableofcontents

\allowdisplaybreaks

\section{Introduction}
Matrix completion has become one of the more active fields in signal 
processing, enjoying numerous applications to data mining and machine learning 
tasks. The matrix completion problem is one where we are allowed to observe a 
small percentage of the entries in a data matrix $\bm{M}$ and from these known 
entries, we must infer the values of the remaining entries. This problem is 
severely ill-posed, particularly so in the high dimensional regime. To this end, 
one must typically assume some sort of low complexity prior on $\bm{M}$, i.e. 
$\bm{M}$ is a low rank matrix or is well approximated by a low rank matrix. 
Using this hypothesis a wide range of theoretical guarantees have been 
established for matrix completion \cite{Cai, CandesRecht, CandesTao, Gross, 
Jain, Oh, NW, Recht}. As noted in \cite{LeverageScoresMCP}, these articles share 
a common thread that the recovery guarantees all require that:
\begin{itemize}
 \item The method of sampling the data matrix $\bm{M}$ must be done in a uniformly random fashion,
 \item And that the low-rank matrix $\bm{M}$ must satisfy a so-called ``incoherence'' property, which roughly means that the distribution of the entries of the matrix must have some form of uniform regularity (\emph{thereby allowing the uniform sampling strategy to be effective}). 
\end{itemize}
In \cite{LeverageScoresMCP} it is observed that although the aforementioned articles differ in optimization techniques, ranging from convex relaxation via nuclear norm minimization \cite{CandesRecht}, non-convex alternating minimization \cite{Jain} and iterative soft thresholding \cite{Cai}, all of these algorithms have exact recovery guarantees using as few as $\Theta(nr \log n)$ observed elements for a square $n \times n$ matrix of rank-$r$. 

One of the central issues in matrix completion is the relationship between the distribution of a matrix's entries and the sampling distribution being employed. For instance, if a matrix is highly incoherent, it has much of its Frobenius norm energy spread throughout its entries in a relatively uniform fashion. To this end, taking a uniformly random sample of this matrix's entries will be a sufficient enough representation to allow for exact recovery. However, if a matrix is highly coherent, in other words, it has much of its Frobenius norm concentrated in a relatively sparse number of its entries, intuitively we understand that a uniform sampling strategy will not yield a sufficiently representative sample to allow for exact recovery. 

Up until recently, the exact nature of this relationship between the 
$\bm{M}$ and the sampling distribution $p$ has not been quantified beyond the 
uniform sampling case. In \cite{LeverageScoresMCP} we see this aforementioned 
relationship quantified. For the purposes and aims of this 
article, we focus on two particular results established in 
\cite{LeverageScoresMCP}:
\begin{itemize}
 \item If the sampling distribution $\bm{p}$ is proportional to the sum of the underlying matrix's \emph{leverage scores}, then any arbitrary $n \times n$ rank-$r$ matrix can be recovered from $\Theta(nr \log^2 n)$ observed entries with high probability. The exact recovery guarantee is for the nuclear norm minimization algorithm \cite{Fazel}. 
 \item Given a set of weights $\bm{R},\bm{C}$, a sufficiency condition on the 
sampling distribution $\bm{p}$ is established. In particular, if the sampling 
distribution $\bm{p}$ is proportional to a sum of these $\bm{R},\bm{C}$ weights, 
then exact recovery guarantees are derived for \emph{weighted nuclear norm 
minimization} (the particular form of weighted nuclear norm minimization 
objective was first posed in \cite{Srebro1,Srebro2}). Moreover, the benefit of 
weighted nuclear norm minimization vs. unweighted nuclear norm minimization is 
quantified with a specific set of weights $\bm{R},\bm{C}$ which are chosen in 
terms of the sampling distribution $\bm{p}$. 
\end{itemize}

We are primarily interested in the second result on weighted nuclear norm 
minimization. We will explore the nature of the relationship between the weights 
$\bm{R},\bm{C}$ and the empirical sampling distribution $\bm{\hat{p}}$ as 
opposed to the true sampling distribution $\bm{p}$. As previously noted, 
\cite{LeverageScoresMCP} established the efficacy of weights $\bm{R},\bm{C}$ 
chosen in a specific fashion in terms of the sampling distribution $\bm{p}$. 
However, we are interested in a setting where the sampling distribution $\bm{p}$ 
is not known to us and no prior knowledge of $\bm{p}$ is available. In this 
article, we make the following contributions:
\begin{enumerate}
 \item We extend the sufficiency condition from \cite{LeverageScoresMCP} to the case when the weights $\bm{R},\bm{C}$ are functions of the empirical sampling distribution $\bm{\hat{p}}$ for the exact recovery of $\bm{M}$ using weighted nuclear norm minimization. 
 \item We show that a specific choice of 
weights $\bm{R},\bm{C}$ as functions of $\bm{\hat{p}}$ produces a similar 
quantifiable relaxation in exact recovery conditions for weighted nuclear 
norm minimization vs. unweighted nuclear norm minimization. 
 \item We numerically demonstrate the healthy robustness of the weighted nuclear 
norm minimization to the choice of the weights $\bm{R},\bm{C}$, hearkening back 
to the previous work in non-uniform sampling and weighted matrix completion 
\cite{Srebro1, Srebro2}. We also demonstrate the superiority of weighted nuclear 
norm minimization over unweighted nuclear norm minimization in the non-uniform 
sampling regime. 
\end{enumerate}
To obtain the above two theoretical guarantees we will use a large deviation and a concentration of measure bound from \cite{Hoeffding} to derive sufficient conditions as to when we may use the empirical sampling distribution $\bm{\hat{p}}$ as an effective proxy for the true sampling distribution $\bm{p}$.
The remainder of the article is organized as follow: in Section 2 we state our main results, in Section 3 we develop all the empirical estimation guarantees required to establish the matrix completion guarantees, in Section 4 we establish our matrix completion guarantees and in Section 5 we present our numerical simulations. 

We use the notation that $a \wedge b := \min(a,b)$ and $a \vee b := 
\max(a,b)$ throughout the article. 

\section{Main Results}

Numerous matrix completion results \cite{CandesRecht, CandesTao, Recht, Fazel} have established the effectiveness of using nuclear norm minimization: 
\begin{equation}\label{nucnorm}
\min_{X \in \mathbb{R}^{n_1 \times n_2}} \|\bm{X}\|_* \textrm{ subject to } X_{ij} = M_{ij} \textrm{ for } (i,j) \in \Omega,
\end{equation}
as a method of performing matrix completion, or in general low rank matrix recovery tasks. However, all of these results may be classified as being in the uniform sampling regime. To this end, recently \cite{LeverageScoresMCP} established that \eqref{nucnorm} can exactly recover an $n \times n$ square matrix $\bm{M}$ of rank-$r$ from $\Theta(nr \log^2 n)$ samples as long as the sampling distribution $\bm{p}$ and $\bm{M}$'s row and column leverage scores $\{ \mu_i(\bm{M}), \nu_j(\bm{M})  \}_{ i,j = 1}^n$ respectively, satisfies the following inequality:
\begin{equation}\label{wardinequality}
 p_{ij} \geq \min\left( c_0 \frac{ (\mu_i(\bm{M})+ \nu_j(\bm{M}))r \log^2(2n)    }{n}, 1     \right) \textrm{   for all } (i,j) \in [n]\times[n],
\end{equation}
for some universal constant $c_0$. With \eqref{wardinequality} the 
quantitative nature between the degree of non-uniformity of the sampling 
distribution $\bm{p}$ and the corresponding coherence statistics of the matrix 
$\bm{M}$ has been established. 

Consider now a different scenario, one in which the sampling distribution $\bm{p}$ and the underlying matrix's leverage scores $\{\mu_i(\bm{M})\}_{i = 1}^{n_1}, \{\nu_j(\bm{M})\}_{j = 1}^{n_2}$ do not align according to \eqref{wardinequality}. One technique to remedy this situation is to design a transformation $\bm{M} \mapsto \bar{\bm{M}}$ so that \emph{we may adjust the leverage scores to align with the sampling distribution $\bm{p}$}. Following \cite{Srebro2, Srebro1} we choose weights of the form $\bm{R} := \mathrm{diag}(R_1, \dots, R_{n_1}) \in \mathbb{R}^{n_1 \times n_1}, \bm{C} := \mathrm{diag}(C_1, \dots, C_{n_2}) \in \mathbb{R}^{n_2 \times n_2}$. Using these parameterized weights, we will use $\bm{M} \mapsto \bm{RMC}$ as our transformation which will adjust leverage scores of $\bm{M}$. In \cite{Srebro1} a weighted nuclear norm objective was proposed. Following \cite{Srebro2, LeverageScoresMCP}, we will be considering the following weighted nuclear norm optimization problem:
\begin{equation}\label{weightedNN}
 \bar{\bm{M}} = \argmin_{\bm{X} \in \mathbb{R}^{n_1 \times n_2}} \|\bm{RXC}\|_* \textrm{ subject to } X_{ij} = M_{ij}, \textrm{ for } (i,j) \in \Omega. 
\end{equation}
In \cite{LeverageScoresMCP} exact recovery guarantees for \eqref{weightedNN}
were established for weights $\bm{R}, \bm{C}$ which were defined in terms of the 
true sampling distribution $\bm{p}$, which we state for the square $n \times n$ case:
\begin{theorem}\label{weightedMCPresult} 
Let $\bm{M} = (M_{ij})$ be an $n \times n$ matrix of rank-$r$, and suppose 
that its elements $M_{ij}$ are observed only over a subset of elements $\Omega 
\subset [n] \times [n]$.
Without loss of generality, assume $R_1 \leq R_2 \leq \cdots \leq R_n$ and 
$C_1 \leq C_2 \leq \cdots \leq C_n$. There exists a universal constant 
$c_0$ 
such that $\bm{M}$ is the unique optimum to \eqref{weightedNN} with 
probability 
at least $1-5(2n)^{-10}$ provided that for all $(i,j) \in 
[n]\times[n], 
p_{ij} \geq n^{-10}$ and:
\begin{equation}\label{weightstransformed}
 p_{ij} \geq c_0 \left(   \frac{R_i^2}{\sum_{i' = 1}^{ \lfloor(n/(\mu_0 
r)\rfloor  }R_{i'}^2   } + \frac{C_j^2}{\sum_{j' = 1}^{ \lfloor(n/(\mu_0 
r)\rfloor  }C_{j'}^2   }  \right) \log^2 (2n).
\end{equation}
\end{theorem} 
Note that for monotonically increasing weights $\bm{R}, \bm{C}$ the 
corresponding support sets $\mathcal{S}_r, \mathcal{S}_c$ are merely the first 
$\lfloor n/(\mu_0 r)   \rfloor$ indices, respectively. 

For the remainder of the article, we shall assume that our sampling distribution $\bm{p}$ has a product form $p_{ij} = p^r_i p^c_j$ for all $(i,j) \in [n_1] \times [n_2]$.  Furthermore, we will consider the following \emph{two-stage sampling model}:
\begin{itemize}
 \item Stage 1 (Empirical Sampling Distribution): We sample the distribution $\bm{p}$ with $m$ times independently with replacement, but the corresponding entries of the data matrix $\bm{M}$ are not revealed to us. In other words, we are \emph{sampling the sampling distribution}, but not the underlying matrix $\bm{M}$. 
 \item Stage 2 (Sampling the Matrix): We then, independent of the first stage, sample the matrix $\bm{M}$ using the independent Bernoulli model for each entry $(i,j) \in [n_1] \times [n_2]$.  
\end{itemize}
Note that this two stage sampling models allows one to sample the sampling distribution $\bm{p}$ without revealing the entries of $\bm{M}$. In this manner we may design weights $\bm{R}, \bm{C}$ which depend on the empirical sampling distribution $\bm{\hat{p}}$ and obtain matrix completion guarantees for these weights in the usual (stage two) independent Bernoulli sampling model that has been typically used in the matrix completion literature. 

In this article we present stage one sampling bounds which will allow 
$\bm{\hat{p}}$ to be used as an empirical proxy for $\bm{p}$ to design weights 
$\bm{R}, \bm{C}$ for \eqref{weightedNN} and obtain exact recovery with high 
probability. To this end, we establish the following two empirical estimation 
lemmas, which will serve as the foundation to our matrix completion guarantees. 
The first is a \emph{one sided large deviation bound}:
\begin{lemma}\label{onesidedphatboundhoeffding}
Let $\bm{p}$ denote a probability mass function on $[n_1] \times [n_2]$ and suppose $\bm{p}$ has a product form, i.e. for all $(i,j) \in [n_1] \times [n_2]: p_{ij} = p^r_i p^c_j$ for $\bm{p^r}, \bm{p^c}$ probability mass functions on $[n_1], [n_2]$, respectively. Let $X_1, \dots, X_m \stackrel{i.i.d}{\sim} \bm{p}$ be a sequence of $m$ i.i.d samples. For any $\alpha \in (0, ( \min_{i \in [n_1]} p^r_i \vee \min_{j \in [n_2]} p^c_j)^{-1}    )$ and $\epsilon \in (0,1)$, if the number of samples $m$ is chosen such that:
\begin{equation} \label{onesidedmbound}
m =  \frac{1}{2}\left(\alpha \min_{i \in [n_1]} p^r_i \wedge \min_{j \in [n_2]} p^c_j \right)^{-2}\log(\epsilon^{-1}(n_1+n_2)  ) ,
\end{equation}
then with probability at least $1-\epsilon$ we have that for all $(i,j) \in [n_1] \times [n_2]$:
\begin{equation}\label{onesidedphatbound}
p_{ij} \geq \frac{1}{(1+\alpha)^2} \hat{p}_{ij}.
\end{equation}
\end{lemma}

We also establish the following \emph{two sided empirical bound} for the 
estimation of product distributions:
\begin{lemma} \label{twosidedphatboundhoeffding} 
Let $\bm{p}$ denote a probability mass function on $[n_1] \times [n_2]$ and suppose $\bm{p}$ has a product form, i.e. for all $(i,j) \in [n_1] \times [n_2]: p_{ij} = p^r_i p^c_j$ for $\bm{p^r}, \bm{p^c}$ probability mass functions on $[n_1], [n_2]$, respectively. Let $X_1, \dots, X_m \stackrel{i.i.d}{\sim} \bm{p}$ be a sequence of $m$ i.i.d samples. For any  $\alpha \in (0, ( \min_{i \in [n_1]} p^r_i \vee \min_{j \in [n_2]} p^c_j)^{-1}    )$ and $\epsilon \in (0,1)$, if the number of samples $m$ is chosen such that:
\begin{equation} \label{twosidedmbound}
m = \frac{1}{2}\left(\alpha \min_{i \in [n_1]} p^r_i \wedge \min_{j \in [n_2]} p^c_j  \right)^{-2}\log(2 \epsilon^{-1}(n_1+n_2) ) ,
\end{equation}
then with probability at least $1-\epsilon$ we have that for all $(i,j) \in [n_1] \times [n_2]$:
\begin{equation}\label{twosidedphatbound}
  \frac{1}{(1+\alpha)^2} \hat{p}_{ij} \leq p_{ij} \leq   \frac{1}{(1-\alpha)^2} \hat{p}_{ij}.
\end{equation}
\end{lemma}

Note that Lemmas \ref{onesidedphatboundhoeffding} and \ref{twosidedphatboundhoeffding} are general results for the empirical 
estimation of any distribution $\bm{p}$ over $[n_1] \times [n_2]$ which has a product form. Recall that the 
sampling model employed in \cite{LeverageScoresMCP} is a sequence of $n_1\cdot 
n_2$ independent Bernoulli random variables, with each Bernoulli random 
variable having success probability $p_{ij}$ for $(i,j) \in [n_1] \times 
[n_2]$. Therefore, $\bm{p}$ may not be a probability matrix on $[n_1]\times[n_2]$ as it may not 
sum to 1. To this end, we note that when we sample $\bm{p}$, we are really 
sampling the normalized matrix $\frac{1}{\sum_{i,j} p_{ij}} \bm{p}$. So our 
empirical estimator $\bm{\hat{p}}$ is estimating the normalized probability matrix 
$\frac{1}{\sum_{i,j} p_{ij}} \bm{p}$ and not $\bm{p}$ itself. Therefore, in 
order to apply the above lemmas we must account for this normalization 
constant. 

Using the above, we will obtain two weighted matrix completion guarantees. For 
simplicity, we will prove all our results for the case when $\bm{M}$ is a 
square $n \times n$ matrix. The first guarantee will be a sufficiency condition 
for the weights $\bm{R}, \bm{C}$ in terms of the empirical estimator 
$\hat{\bm{p}}$ which will ensure exact recovery by weighted nuclear norm 
minimization with high probability:

\begin{theorem}\label{sufficientwmcp}
 Let $\bm{M} = (M_{ij})$ be an $n \times n$ matrix of rank-$r$, and suppose that its elements $M_{ij}$ are observed only over a subset of elements $\Omega \subset [n] \times [n]$, Let $\epsilon \in (0,1)$ be arbitrary. Suppose that there exists an  $\alpha \in (0, ( \min_{i \in [n]} p^r_i/(\sum_{i \in [n]} p^r_i) \vee \min_{j \in [n]} p^c_j/(\sum_{j \in [n]} p^c_j))^{-1}    )$ and some universal constant $c_0$ such that for all indices $(i,j) \in [n] \times [n]$ the weights $\bm{R}, \bm{C}$ satisfy the following inequalities:
\begin{equation}\label{onesidedtheoreminequality}
 \hat{p}_{ij} \geq \frac{(1+\alpha)^2}{\sum_{ij} p_{ij}}c_0 \left(  \frac{R_i^2}{\sum_{i' \in\mathcal{S}_r} R_{i'}^2   } +  \frac{C_j^2}{\sum_{j' \in\mathcal{S}_c} C_{j'}^2   }    \right) \log^2 (2n),
\end{equation}
where $\mathcal{S}_r, \mathcal{S}_c$ denote the $\lfloor n/(\mu_0 r)   \rfloor$ entries of least magnitude of $\bm{R}, \bm{C}$, respectively. If the number of stage one samples $m$ is chosen such that:
$$
m =  \frac{1}{2} \left(\alpha \min_{i \in [n]} \frac{p^r_i}{\sum_{i = 1}^{n} p^r_i} \wedge \min_{j \in [n]} \frac{p^c_j}{\sum_{j = 1}^{n} p^c_j}   \right)^{-2} \log(2\epsilon^{-1}n)
$$
and if for all $(i,j) \in [n]\times[n], p_{ij} \geq n^{-10}$, then with probability at least $(1-5(2n)^{-10})(1-\epsilon), \bm{M}$ is unique optimum to \eqref{weightedNN}, where $\Omega$ is obtained via the usual (stage two) independent, entry-wise Bernoulli sampling of $\bm{M}$. 
\end{theorem}
Our second weighted matrix completion guarantee will be for the exact recovery properties of a set weights $\bm{R}, \bm{C}$ explicitly defined in terms of the empirical distribution $\bm{\hat{p}}$:
\begin{theorem}\label{specificmcp}Let $\bm{M}$ be a square $n\times n$ 
rank-$r$ matrix with coherence $\mu_0$. Consider the 
weights defined by:
\begin{align}
\label{empweightsrow} R_i &= \sqrt{ \frac{1}{n} \hat{p}^r_i \sum_{j' \in \mathcal{S}_c} \hat{p}^c_{j'}    } \textrm{ for } i = 
1,\dots, n,
\\ \label{empweightscol} C_j &= \sqrt{ \frac{1}{n} \hat{p}^c_j \sum_{i' \in \mathcal{S}_r} 
\hat{p}^r_{i'}    } \textrm{ for } j = 1,\dots, n,
\end{align}
where $\mathcal{S}_r, \mathcal{S}_c$ denote the $\lfloor n/(\mu_0 r)   \rfloor$ entries of $\bm{\hat{p}^r}, \bm{\hat{p}^c}$ of least magnitude, respectively. 
Suppose that there exists an  $\alpha \in (0, ( \min_{i \in [n]} p^r_i/(\sum_{i \in [n]} p^r_i) \vee \min_{j \in [n]} p^c_j/(\sum_{j \in [n]} p^c_j))^{-1}    )$ such that the 
(unnormalized) matrix $\bm{p}$ satisfies for all $(i,j) \in [n] \times [n]$ and 
the sets $\mathcal{S}_r^*, \mathcal{S}_c^*$ which denote the $\lfloor n/(\mu_0 
r)   \rfloor$ entries of $\bm{p^r}, \bm{p^c}$ of least magnitude, respectively 
satisfies the following:
\begin{align}\label{lessrestrictivea}
p^c_j \sum_{i'\in \mathcal{S}_r^*} p^r_{i'}&\geq c_0 
\frac{2(1+\alpha)^2}{(1-\alpha)^2} \log^2(2n),
\\ \label{lessrestrictiveb} p^r_i \sum_{j'\in \mathcal{S}_c^*} p^c_{j'} &\geq 
c_0 \frac{2(1+\alpha)^2}{(1-\alpha)^2} \log^2(2n).
\end{align}
If the number of stage one samples $m$ is chosen such that:
$$
m =  \frac{1}{2} \left(\alpha \min_{i \in [n]} \frac{p^r_i}{\sum_{i = 1}^{n} p^r_i} \wedge \min_{j \in [n]} \frac{p^c_j}{\sum_{j = 1}^{n} p^c_j}   \right)^{-2} \log(4\epsilon^{-1}n), 
$$ then with probability at least $(1-5(2n)^{-10})(1-\epsilon), \bm{M}$ is unique optimum to \eqref{weightedNN}, where $\Omega$ is obtained via the usual (stage two) independent, entry-wise Bernoulli sampling of $\bm{M}$. 
\end{theorem}
\emph{Note:} Unweighted nuclear norm minimization attains exact recovery under 
the condition that for all $(i,j) \in [n]\times[n]$:
\begin{equation}\label{unweightednuclearnormsufficient}
p^r_i p^c_j \gtrsim \frac{\mu_0 r}{n} \log^2 (2n). 
\end{equation}
\noindent However as Theorem \ref{specificmcp} establishes, weighted nuclear 
norm minimization with choice of weights \eqref{empweightsrow} and 
\eqref{empweightscol} attains exact recovery subject to the less restrictive 
sufficient recovery condition that:
\begin{align*}
p^c_j \sum_{i' \in \mathcal{S}_r^*} p^r_{i'}&\gtrsim
\log^2 (2n),
\\  p^r_i \sum_{j'\in \mathcal{S}_c^*} p^c_{j'} &\gtrsim 
\log^2 (2n).
\end{align*}
This is precisely the condition from
\cite{LeverageScoresMCP}. 
\section{Empirical Estimation} 
We consider probability mass functions $\bm{p}$ on $[n_1]\times[n_2]$ which have a product form $p_{ij} = p^r_i p^c_j$ for $(i,j) \in [n_1] \times[n_2]$. We will sample this distribution with replacement $m$ times. The $X_1, \dots, X_m \stackrel{i.i.d}{\sim} \bm{p}$ samples are row and column pairs, i.e. $X_k \in [n_1] \times [n_2]$ for each $k = 1, \dots, m$. We may define the \emph{row and column empirical estimators}:
\begin{definition}
The row and column empirical estimators $\bm{\hat{p}^r}, \bm{\hat{p}^c}$, respectively are defined as:
\begin{align}
\label{rowempiricalestimator} \hat{p}^r_i &:= \frac{1}{m} \sum_{k = 1}^m \delta^r_i(X_k), \textrm{ for } i \in [n_1],
\\ \label{colempiricalestimator}  \hat{p}^c_j &:= \frac{1}{m} \sum_{k = 1}^m \delta^c_j(X_k), \textrm{ for } j \in [n_2],
\end{align}
where for any $X_k$:
\begin{align*}
\delta^r_i(X_k) &= \begin{cases}
                  1 & \textrm{ if } X_k \textrm{ is from row } i,
                  \\ 0 & \textrm{ otherwise}. 
                  \end{cases} 
\\
\delta^c_j(X_k)  &= \begin{cases}
                  1 & \textrm{ if } X_k \textrm{ is from column } j,
                  \\ 0 & \textrm{ otherwise}. 
                  \end{cases} 
\end{align*}
\end{definition}
For the remainder of the article, we will allow $\bm{\hat{p}}$ denote the empirical product estimate, i.e. $\bm{\hat{p}} = \bm{\hat{p}^r} \bm{\hat{p}^c}$. 

Observe that in \eqref{rowempiricalestimator} and \eqref{colempiricalestimator} each component of our row and column empirical estimators involve a sum of independent, bounded in $[0,1]$ random variables as $\delta^r_i(X_k), \delta^c_j(X_k) \in \{0,1\}$ for any $(i,j,k) \in [n_1] \times [n_2] \times [m]$. In this situation, we may use \emph{Hoeffding's inequalities} \cite{Hoeffding} to obtain some probabilistic approximation guarantees. For our purposes, we will be using two forms of Hoeffding's inequalities: a one sided large deviation bound and a two sided concentration of measure bound. 

\begin{theorem}\label{hoeffding} (Hoeffding Inequalities) Let $Z_1, \dots, Z_m$ be independent random variables such that each $Z_i \in [a_i, b_i]$ with probability 1. Let $S_m = \sum_{i = 1}^m Z_i$. Then for any $t > 0$ we have:
\begin{align}
\label{onesidedhoeffdinginequality} \Pr[ S_m - \E[S_m] \geq t] &\leq \exp\left( - \frac{2t^2}{ \sum_{i = 1}^m (b_i-a_i)^2  }     \right),
\\ \label{twosidedhoeffdinginequality} \Pr[ |S_m - \E[S_m]| \geq t] &\leq 2\exp\left( - \frac{2t^2}{ \sum_{i = 1}^m (b_i-a_i)^2  }     \right). 
\end{align}
\end{theorem}

For any $i \in [n_1]$, we may define $m$ random variables $Z_{i,k} := 
\delta^r_i(X_k)$ for $k = 1,\dots, m$. Note that each random variable $Z_{i,k}$ 
only takes values in $\{0,1\}$ and thus is bounded in $[0,1]$ with probability 
1. As each $X_k$ is merely a row and column index, and each $\delta^r_i, 
\delta^c_j$ are row and column indicator functions, we have that any set of the 
$Z_{i,k}$'s (and similarly for the column case) is an independent set of random 
variables. Therefore the hypotheses of Theorem \ref{hoeffding} are satisfied. 
For each $i \in [n_1]$ we may define the sum $S^r_{i,m} := \sum_{ k = 1}^m 
Z_{i,k}$. Each $S^r_{i,m}$ has expected value $\E[S^r_{i,m}] = mp^r_i$. 
Analogous results hold for the column case. With the above pair of Hoeffding 
inequalities in hand, we are now ready to establish our main lemmas. For the 
proof of Lemma \ref{onesidedphatboundhoeffding} we will apply 
\eqref{onesidedhoeffdinginequality} and for the proof of Lemma 
\ref{twosidedphatboundhoeffding} we will apply 
\eqref{twosidedhoeffdinginequality}.

\subsection{Proof Lemma \ref{onesidedphatboundhoeffding}} 
\begin{proof}
We start our proof by analyzing empirical estimation of the row distribution; the analysis for the column distribution will be identical. For any $i \in [n_1], \alpha > 0$, choosing $t = \alpha \min_{i \in [n_1]} p^r_i$, by \eqref{onesidedhoeffdinginequality} we have that:
\begin{equation}
\label{onesidedrowtbound} \Pr[\hat{p}^r_i - p^r_i \geq \alpha \min_{i \in [n_1]} p^r_i] \leq  \exp(-2(\alpha \min_{i \in [n_1]} p^r_i)^2m).
\end{equation}
We may repeat the analysis for the column case, where we choose $t = \alpha \min_{j \in [n_2]} p^c_j$, then analogously:
\begin{equation}\label{onesidedcoltbound}
\Pr[\hat{p}^c_j - p^c_j \geq \alpha \min_{j \in [n_2]} p^c_j]  \leq \exp(-2(\alpha \min_{j \in [n_2]} p^c_j)^2m).
\end{equation}
For any $i \in [n_1]$ let $E^r_i$ denote the event that $\hat{p}^r_i - p^r_i \geq \alpha \min_{i \in [n_1]} p^r_i$ and for any $j \in [n_2]$ let $E^c_j$ denote the event that $\hat{p}^c_j - p^c_j \geq \alpha \min_{j \in [n_2]} p^c_j$. 

We must choose $\alpha > 0$ such that the bounds in \eqref{onesidedrowtbound}, \eqref{onesidedcoltbound} are nontrivial. In particular, any two probability vectors cannot have their components differ by more than 1. Therefore, we require that $\alpha$ satisfies:
$$
\alpha \min_{i \in [n_1]} p^r_i \leq 1 \textrm{ and } \alpha\min_{j \in [n_2]} p^c_j \leq 1.
$$
To this end it suffices to choose $\alpha \in (0, ( \min_{i \in [n_1]} p^r_i \vee \min_{j \in [n_2]} p^c_j)^{-1}    )$. 

By \eqref{onesidedrowtbound}, \eqref{onesidedcoltbound} and the Union Bound we have that:
\begin{align}
\Pr\left[\textrm{For some } (i,j)\textrm{ the event } E^r_i \textrm{ or } E^c_j \textrm{ occurs}  \right] &\leq \left( n_1\exp(-2(\alpha \min_{i \in [n_1]} p^r_i)^2m) + n_2\exp(-2(\alpha \min_{j \in [n_2]} p^c_j)^2m    \right) \nonumber
\\ \label{onesidedunionbound} &\leq (n_1 +n_2)\exp(-2(\alpha \min_{i \in [n_1]} p^r_i \wedge \min_{j \in [n_2]} p^c_j)^2m).
\end{align}

Observe that \eqref{onesidedunionbound} immediately yields that with probability at least $1 - (n_1 + n_2)\exp(-2(\alpha\min_{i \in [n_1]} p^r_i \wedge \min_{j \in [n_2]} p^c_j)^2m)$ for any $(i,j) \in [n_1] \times [n_2]$ we have that the following bounds hold:
\begin{align}
\label{onesidedbasicbound1} \hat{p}^r_i - p^r_i &\leq \alpha \min_{i \in [n_1]} p^r_i,
\\ \label{onesidedbasicbound2} \hat{p}^c_j - p^c_j &\leq \alpha \min_{j \in [n_2]} p^c_j.
\end{align}
Therefore with probability at least $1 - (n_1 + n_2)\exp(-2(\alpha \min p_{i,j})^2m)$ we may conclude that for all $(i,j) \in [n_1] \times [n_2]$ the following bound is true:
\begin{equation}\label{onesidedmainbound}
 p_{ij} \geq \frac{1}{(1+\alpha)^2} \hat{p}_{ij}.
\end{equation}

For any $\epsilon \in (0,1)$ choosing $m$ such that:
\begin{equation}\label{monesided}
m = \frac{ \log((n_1+n_2) \epsilon^{-1} )   }{2  \left(\alpha\min_{i \in [n_1]} p^r_i \wedge \min_{j \in [n_2]} p^c_j\right)^2},
\end{equation}
guarantees that \eqref{onesidedmainbound} holds with probability at least $(1-\epsilon)$ and the proof is complete. \end{proof}

\subsection{Proof of Lemma \ref{twosidedphatboundhoeffding}} 
\begin{proof} 
The proof of Lemma \ref{twosidedphatboundhoeffding} is similar to the previous proof but we include the full proof for completeness. 
We start our proof by analyzing empirical estimation of the row distribution; the analysis for the column distribution will be identical. Following the previous section we restrict ourselves to choose $\alpha \in (0, ( \min_{i \in [n_1]} p^r_i \vee \min_{j \in [n_2]} p^c_j)^{-1}   )$. For any $i \in [n_1]$ choosing $t = \alpha \min_{i \in [n_1]} p^r_i$, by \eqref{twosidedhoeffdinginequality} we have that:
\begin{equation}
\label{twosidedrowtbound} \Pr[|\hat{p}^r_i - p^r_i| \geq \alpha \min_{i \in [n_1]} p^r_i] \leq  2\exp(-2(\alpha \min_{i \in [n_1]} p^r_i)^2m).
\end{equation}

We may repeat the analysis for the column case, where we choose $t = \alpha \min_{j \in [n_2]} p^c_j$, then analogously:
\begin{equation}\label{twosidedcoltbound}
\Pr[|\hat{p}^c_j - p^c_j| \geq \alpha \min_{j \in [n_2]} p^c_j]  \leq 2\exp(-2(\alpha \min_{j \in [n_2]} p^c_j)^2m).
\end{equation}

For any $i \in [n_1]$ let $E^r_i$ denote the event that $|\hat{p}^r_i - p^r_i| \geq \alpha \min_{i \in [n_1]} p^r_i$ and for any $j \in [n_2]$ let $E^c_j$ denote the event that $|\hat{p}^c_j - p^c_j| \geq \alpha \min_{j \in [n_2]} p^c_j$. By \eqref{twosidedrowtbound}, \eqref{twosidedcoltbound} and the Union Bound we have that:
\begin{align}
\Pr\left[\textrm{For some } (i,j)\textrm{ the event } E^r_i \textrm{ or } E^c_j \textrm{ occurs}  \right] &\leq 2\left( n_1\exp(-2(\alpha \min_{i \in [n_1]} p^r_i)^2m) + n_2\exp(-2(\alpha \min_{j \in [n_2]} p^c_j)^2m    \right) \nonumber
\\ \label{unionbound} &\leq 2(n_1 +n_2)\exp(-2(\alpha \min_{i \in [n_1]} p^r_i \wedge \min_{j \in [n_2]} p^c_j   )^2m).
\end{align}
Observe that \eqref{unionbound} immediately yields that with probability at least $1 - 2(n_1 + n_2)\exp(-2(\alpha \min_{i \in [n_1]} p^r_i \wedge \min_{j \in [n_2]} p^c_j )^2m)$ for any $(i,j) \in [n_1] \times [n_2]$ we have that the two following bounds hold:
\begin{align}
\label{basicbound1} |\hat{p}^r_i - p^r_i| &\leq \alpha \min_{i \in [n_1]} p^r_i,
\\ \label{basicbound2} |\hat{p}^c_j - p^c_j| &\leq \alpha \min_{j \in [n_2]} p^c_j.
\end{align}

The bound \eqref{basicbound1} is equivalent to the following:
$$
-\alpha \min_{i \in [n_1]} p^r_i \leq \hat{p}^r_i - p^r_i \leq \alpha \min_{i \in [n_1]} p^r_i,
$$
and the above inequality yields that for any $i \in [n_1]$:
\begin{equation}\label{generaltwosidedrowbound}
 \frac{1}{1+\alpha} \hat{p}^r_i \leq p^r_i \leq \frac{1}{1-\alpha} \hat{p}^r_i. 
\end{equation}
Similarly \eqref{basicbound2} implies that for any $j \in [n_2]$:
\begin{equation}\label{generaltwosidedcolbound}
  \frac{1}{1+\alpha} \hat{p}^c_j \leq p^c_j \leq \frac{1}{1-\alpha} \hat{p}^c_j. 
\end{equation}

Combining \eqref{generaltwosidedrowbound} and \eqref{generaltwosidedcolbound}, we have that with probability at least $1 - 2(n_1 + n_2)\exp(-2(\alpha \min_{i \in [n_1]} p^r_i \wedge \min_{j \in [n_2]} p^c_j )^2m)$ that:
\begin{equation}\label{fulltwosidedhoeffding}
 \frac{1}{(1+\alpha)^2} \bm{\hat{p}} \leq \bm{p} \leq  \frac{1}{(1-\alpha)^2} \bm{\hat{p}}.
\end{equation}
For any $\epsilon \in (0,1)$ note that if we choose:
\begin{equation}\label{mtwosided}
m = \frac{ \log(2(n_1+n_2) \epsilon^{-1} )   }{2 (\alpha \min_{i \in [n_1]} p^r_i \wedge \min_{j \in [n_2]} p^c_j )^2 },
\end{equation}
then \eqref{fulltwosidedhoeffding} holds with probability at least $1-\epsilon$ and the proof is complete. \end{proof}

\section{Matrix Completion Guarantees}
With Lemma \ref{onesidedphatboundhoeffding} in hand, we are prepared to prove Theorem \ref{sufficientwmcp} in Section 4.1. In Section 4.2, using Lemma \ref{twosidedphatboundhoeffding} we will prove Theorem \ref{specificmcp} which quantifies the relaxation of the condition for which \eqref{weightedNN} succeeds in obtaining  exact recovery using the empirically learned weights when compared to unweighted nuclear norm minimization. 

\subsection{Proof of Theorem \ref{sufficientwmcp}}
\begin{proof}
For any $\alpha \in (0, ( \min_{i \in [n]} p^r_i/(\sum_{i = 1}^n p^r_i) \vee \min_{j \in [n]} p^c_j/(\sum_{j = 1}^n p^c_j))^{-1}    )$ and $\epsilon \in (0,1)$ if we choose 
$$m =  \frac{1}{2} \left(\alpha \min_{i \in [n]} \frac{p^r_i}{\sum_{i = 1}^{n} p^r_i} \wedge \min_{j \in [n]} \frac{p^c_j}{\sum_{j = 1}^{n} p^c_j}   \right)^{-2} \log(2\epsilon^{-1}n)$$
by Lemma \ref{onesidedphatboundhoeffding} we have that with probability at least $(1-\epsilon)$ for any $(i,j) \in [n] \times [n]$:
\begin{equation}\label{theoremonesideda}
 \frac{p_{ij}}{\sum_{ij} p_{ij} } \geq \frac{1}{(1+\alpha)^2} \hat{p}_{ij}. 
\end{equation}
Observe that if the weights $\bm{R}, \bm{C}$ satisfy \eqref{onesidedtheoreminequality} for $\alpha$, we have that:
\begin{align}
p_{ij} &\geq \frac{\sum_{ij} p_{ij} }{(1+\alpha)^2} \hat{p}_{ij}
\\ \label{theoremonesidedb} &\geq c_0 \left(  \frac{R_i^2}{\sum_{i' \in\mathcal{S}_r} R_{i'}^2   } +  \frac{C_j^2}{\sum_{j' \in\mathcal{S}_c} C_{j'}^2   }    \right) \log^2 (2n).
\end{align}
By Theorem \ref{weightedMCPresult} \eqref{theoremonesidedb} is sufficient to 
guarantee exact recovery of $\bm{M}$ via \eqref{weightedNN} with probability at 
least $1-5(2n)^{-10}$. As stage one and stage two sampling are independent, we 
conclude that \eqref{weightedNN} attains exact recovery with probability at 
least $(1-5(2n)^{-10})(1-\epsilon)$. 
\end{proof}

\subsection{Weighted Nuclear Norm and Relaxation of Sufficient Recovery 
Conditions}
With Theorem \ref{sufficientwmcp} we established some sufficient 
conditions for the weights $\bm{R}, \bm{C}$ in order for \eqref{weightedNN} to attain exact 
recovery. In this section we will establish exact recovery guarantees for a 
specific set of weights defined in terms of the empirical sampling distribution 
$\bm{\hat{p}}$ and quantify how the exact recovery conditions for \eqref{weightedNN} 
are relaxed relative to unweighted nuclear norm minimization \eqref{nucnorm}. 

\subsubsection{Proof of Theorem \ref{specificmcp}}

\begin{proof}
Choosing the weights $\bm{R}, \bm{C}$ as in \eqref{empweightsrow} and 
\eqref{empweightscol}, observe that for any $(i,j) \in [n]\times[n]$:
\begin{align}
\left(   \frac{R_i^2}{\sum_{i' \in \mathcal{S}_r} R^2_{i'}    
} + \frac{C_j^2}{\sum_{j'\in \mathcal{S}_c} C_{j'}^2   }  
\right) \log^2 (2n) &= \left(   \frac{ \hat{p}^r_i\sum_{j' \in \mathcal{S}_c}\hat{p}^c_{j'} + \hat{p}^c_j\sum_{i' \in \mathcal{S}_r}\hat{p}^r_{i'}}{ \sum_{i',j' \in \mathcal{S}_r, \mathcal{S}_c}\hat{p}^r_{i'}\hat{p}^c_{j'}   }   
\right) 
\log^2(2n) \label{specificweightsintermediate}.
\end{align}

Let $\alpha \in (0, ( \min_{i \in [n]} p^r_i \vee \min_{j \in [n_2]} p^c_j)^{-1}    )$ be such that \eqref{lessrestrictivea} and \eqref{lessrestrictiveb} hold and let  $\epsilon \in (0,1)$ be arbitrary. By Lemma \ref{twosidedphatboundhoeffding} choosing $m$ such that:
$$m =  \frac{1}{2} \left(\alpha \min_{i \in [n]} \frac{p^r_i}{\sum_{i = 1}^{n} p^r_i} \wedge \min_{j \in [n]} \frac{p^c_j}{\sum_{j = 1}^{n} p^c_j}   \right)^{-2} \log(4\epsilon^{-1}n)$$
guarantees that with probability at least $(1-\epsilon)$ that for all indices $(i,j) \in [n]\times[n]$:
\begin{equation}\label{twosidedboundnoprob}
 \frac{1}{(1+\alpha)^2} \hat{p}_{ij} \leq \frac{p_{ij}}{\sum_{i,j} p_{ij}} \leq 
\frac{1}{(1-\alpha)^2} 
\hat{p}_{ij}.
\end{equation}
Applying \eqref{twosidedboundnoprob} to \eqref{specificweightsintermediate} we 
have that for any $(i,j) \in [n] \times [n]$:
\begin{align} 
\left(   \frac{R_i^2}{\sum_{i' \in \mathcal{S}_r} R^2_{i'}    
} + \frac{C_j^2}{\sum_{j'\in \mathcal{S}_c} C_{j'}^2   }  
\right) \log^2 (2n) &= \left(   \frac{ \hat{p}^r_i\sum_{j' \in \mathcal{S}_c}\hat{p}^c_{j'} + \hat{p}^c_j\sum_{i' \in \mathcal{S}_r}\hat{p}^r_{i'}}{ \sum_{i',j' \in \mathcal{S}_r, \mathcal{S}_c}\hat{p}^r_{i'}\hat{p}^c_{j'}   }   
\right) 
\log^2 (2n) \nonumber
\\ &\leq \frac{(1+\alpha)^2}{(1-\alpha)^2}\left(   \frac{p^r_i\sum_{j' \in \mathcal{S}_c}p^c_{j'} + p^c_j\sum_{i'\in \mathcal{S}_r}p^r_{i'}}{ \sum_{i',j'\in \mathcal{S}_r, \mathcal{S}_c}p^r_{i'}p^c_{j'}   }   
\right) 
\log^2 (2n)\nonumber
\\&= \frac{(1+\alpha)^2}{(1-\alpha)^2} \left[  \frac{p^r_i \log^2 (2n)}{  \sum_{i'\in \mathcal{S}_r}p^r_{i'}    }  +  \frac{p^c_j \log^2 (2n)}{  
\sum_{j' \in \mathcal{S}_c}p^c_{j'}    }  \right] \nonumber
\\ \label{minimalprealbound} &\leq \frac{(1+\alpha)^2}{(1-\alpha)^2} \left[  
\frac{p^r_i \log^2 (2n)}{  \sum_{i'\in \mathcal{S}_r^*}p^r_{i'}    }  +  
\frac{p^c_j \log^2 (2n)}{ 
\sum_{j' \in \mathcal{S}_c^*}p^c_{j'}    }  \right] 
\\ \label{doneskies}&\leq \frac{1}{c_0} p_{ij}.
\end{align}
where \eqref{minimalprealbound} follows as the sets $\mathcal{S}_r^*, 
\mathcal{S}_c^*$ serve as a lower bound for the terms $\sum_{i' \in 
\mathcal{S}_r}p^r_{i'}, \sum_{j' \in \mathcal{S}_c}p^c_{j'}$ respectively 
and 
thus inverting they serve as an upper bound and \eqref{doneskies} follows from 
\eqref{lessrestrictivea} and 
\eqref{lessrestrictiveb}. Again by Theorem \ref{weightedMCPresult} we 
immediately see that \eqref{doneskies} is sufficient to guarantee exact recovery 
of $\bm{M}$ via \eqref{weightedNN} with probability at least $1-5(2n)^{-10}$. 
\end{proof}

\section{Numerical Experiments}
\begin{figure}[t!]
\centering
\begin{minipage}[b]{0.45\linewidth}
\includegraphics[scale=.8]{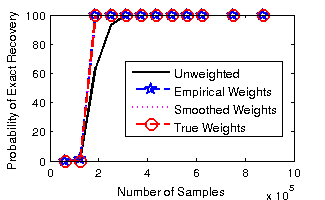}
\caption{Probability of Exact Recovery when the rank is equal to 5.}
\label{fig:rank5}
\end{minipage}
\quad
\begin{minipage}[b]{0.45\linewidth}
\includegraphics[scale=.8]{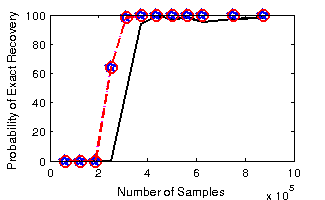}
\caption{Probability of Exact Recovery when the rank is equal to 10.}
\label{fig:rank10}
\end{minipage}
\end{figure}

\begin{figure}[t!]
\centering
\begin{minipage}[b]{0.45\linewidth}
\includegraphics[scale=.8]{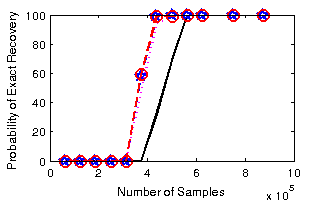}
\caption{Probability of Exact Recovery when the rank is equal to 15.}
\label{fig:rank15}
\end{minipage}
\quad
\begin{minipage}[b]{0.45\linewidth}
\includegraphics[scale=.8]{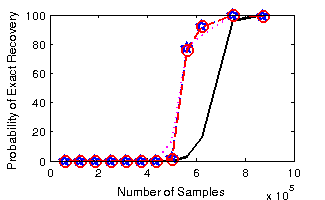}
\caption{Probability of Exact Recovery when the rank is equal to 20.}
\label{fig:rank20}
\end{minipage}
\end{figure}

\begin{figure}[t!]
\centering
\begin{minipage}[b]{0.45\linewidth}
\includegraphics[scale=.8]{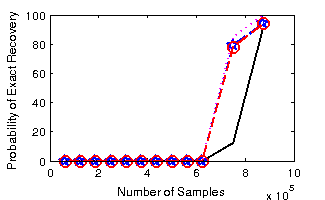}
\caption{Probability of Exact Recovery when the rank is equal to 25.}
\label{fig:rank25}
\end{minipage}
\quad
\begin{minipage}[b]{0.45\linewidth}
\includegraphics[scale=.8]{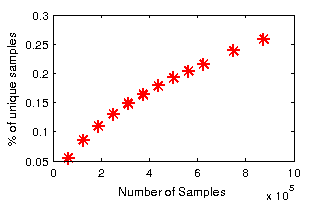}
\caption{Power Law Sampling with replacement rate vs. Percentage of Unique Samples Revealed.}
\label{fig:samplerate}
\end{minipage}
\end{figure}

Here we test the performance of weighted nuclear norm minimization using various weights. We have the following experimental setup: the data matrix $\bm{M}$ is a unit Frobenius norm standard normal Gaussian square matrix of dimension $n = 500$. Our sampling distribution $\bm{p} = \bm{p^r p^c}$ where $\bm{p^r}, \bm{p^c}$ are power law distributed with exponent equal to 1.2. Sampling the distribution $\bm{p}$ at a rate of $m$ times with replacement and we obtain the empirical distribution $\bm{\hat{p}} = \bm{\hat{p}^r \hat{p}^c}$. Using this empirical distribution $\bm{\hat{p}}$ we test nuclear norm minimization using the following weights, as was done in \cite{Srebro2}:
\begin{enumerate}
 \item Unweighted (Uniform Weights): the weights $\bm{R}, \bm{C}$ are equal to the uniform weights. 
 \item True Weighted: the weights $\bm{R}, \bm{C}$ satisfy: $\bm{R} = (\bm{p^r})^{1/2}, \bm{C} = (\bm{p^c})^{1/2}$. 
 \item Empirically Weighted: the weights $\bm{R}, \bm{C}$ satisfy: $\bm{R} = (\bm{\hat{p}^r})^{1/2}, \bm{C} = (\bm{\hat{p}^c})^{1/2}$. 
 \item Empirically Smoothed Weights: the weights $\bm{R}, \bm{C}$ are a linear combination of the empirical weights and the uniform weights. Letting $\bm{1}_n := [1,\dots,1]$ be a vector of length $n$ whose coordinates are all equal to 1, we set $\bm{R} = \frac{1}{2n} (\bm{\hat{p}^r})^{1/2} + \frac{1}{2n} \bm{1}_n$ and $\bm{C} = \frac{1}{2} (\bm{\hat{p}^c})^{1/2} + \frac{1}{2n} \bm{1}_n$, i.e. we put half of the mass on the empirical distribution and remaining half of the mass on the uniform weights. 
\end{enumerate}
We let the rank of $\bm{M}$ be 5, 10, 15, 20, 25 and we choose a range of variable sampling rates. For each rank and sampling rate test configuration we performed 100 trials. We consider exact recovery to be when the output of the weighted nuclear norm $\bar{\bm{M}}$ satisfies: $\| \bm{M} - \bar{\bm{M}} \|_F \leq 10^{-5}$. To execute the weighted nuclear norm minimization program we utilized the Augmented Lagrangian Method \cite{ALM}. We obtained the following phase transition diagrams in Figures \ref{fig:rank5}-\ref{fig:rank25}.  

Note that we do not perform the two stage sampling method. As the power law sampling distribution $\bm{p}$ is non-uniform, even though we may sample at a rate of $m = O(n_1 n_2)$, the rate that the percentage of unique revealed entries of $\bm{M}$ grows is in line with the uniform sampling regime we are accustomed to. In Figure \ref{fig:samplerate} we show how  with the independent sampling with replacement rate $m$ grows with the percentage of unique entries of $\bm{M}$. 

\section{Conclusion}
In this article we extended numerous weighted nuclear norm minimization results from \cite{LeverageScoresMCP}. In particular we extended results where the weights were being defined in relation to the true sampling distribution $\bm{p}$ to the weights being defined in relation to the empirical sampling distribution $\bm{\hat{p}}$. Furthermore, we defined an empirical set of weights and established a quantifiable relaxation of exact recovery conditions for weighted nuclear norm minimization when compared to the unweighted nuclear norm. To achieve these guarantees we used a large deviation bound and a concentration of measure inequality from \cite{Hoeffding}. We showed that weighted nuclear norm minimization is quite robust to the choice of empirically learned weights. Indeed, we used a broad range of empirical weights and saw strikingly similar exact recovery gains over unweighted nuclear norm minimization.

\section{Acknowledgements}
The author would like to acknowledge and thank Rachel Ward for their insight and guidance throughout this project. 
\bibliographystyle{unsrt}  
\bibliography{simplerweightedmcp}

\end{document}